\documentclass{article}
\usepackage{spconf,amsmath,graphicx}
\usepackage{amssymb,amsfonts}
\graphicspath{{figs/}}
\usepackage{booktabs}
\usepackage{etoolbox}
\usepackage{siunitx}
\usepackage{textcomp}
\usepackage{xcolor}
\usepackage{multirow}
\usepackage[caption=false, font=footnotesize]{subfig}
\usepackage[export]{adjustbox}
\usepackage{enumitem}
\usepackage{cite}
\usepackage{mathtools}
\usepackage{xspace}
\usepackage{rotating}
\usepackage{diagbox}
\usepackage{hyperref}

\newcommand{\pyone}[1]{\textcolor{black}{#1}}

\makeatletter
\DeclareRobustCommand\onedot{\futurelet\@let@token\@onedot}
\def\@onedot{\ifx\@let@token.\else.\null\fi\xspace}

\def\eg{{\it e.g}\onedot} 
\def\ie{{\it i.e}\onedot}

\makeatother

\title{FINE-TUNING TEXT-TO-IMAGE DIFFUSION MODELS FOR CLASS-WISE SPURIOUS FEATURE GENERATION}
\name{\begin{tabular}{c}AprilPyone MaungMaung$^{1}$ \quad Huy H. Nguyen$^1$ \quad Hitoshi Kiya$^{2}$ \quad Isao Echizen$^{1,3}$\end{tabular}}

\address{$^{1}$National Institute of Informatics \qquad
 $^{2}$Tokyo Metropolitan University \qquad $^3$University of Tokyo}

\begin{document}
\maketitle
\begin{abstract}
We propose a method for generating spurious features by leveraging large-scale text-to-image diffusion models.
Although the previous work detects spurious features in a large-scale dataset like ImageNet and introduces Spurious ImageNet, we found that not all spurious images are spurious \pyone{across different classifiers.
Although spurious images help measure the reliance of a classifier, filtering many images from the Internet to find more spurious features is time-consuming.}
To this end, we utilize an existing approach of personalizing large-scale text-to-image diffusion models \pyone{with available discovered spurious images} and propose a new spurious feature similarity loss based on neural features of an adversarially robust model.
Precisely, we fine-tune Stable Diffusion with several reference images from Spurious ImageNet with a modified objective incorporating the proposed spurious-feature similarity loss. Experiment results show that our method can generate spurious images that are consistently spurious across different classifiers. Moreover, the generated spurious images are visually similar to reference images from Spurious ImageNet.
\end{abstract}


\section{Introduction}
Deep neural networks (DNNs) have brought state-of-the-art results for visual recognition~\cite{he2016deep}, natural language processing~\cite{vaswani2017attention}, and speech recognition~\cite{graves2013speech}. Although DNNs are ubiquitous, evaluating a DNN is not trivial. The evaluation of DNNs is even more critical when DNNs are deployed in high-stakes settings such as breast cancer screening~\cite{mckinney2020international} and autonomous vehicles~\cite{eykholt2018robust}. Generally, the performance of an image classifier is tested on a fixed (held-out) set. However, this test performance is not guaranteed to reflect the actual performance of a real-world deployment. For example, the ImageNet~\cite{ILSVRC15} test set is simple and easy to classify and does not reflect real-world performance~\cite{recht2019imagenet}. Therefore, image classifiers are generally evaluated against common image corruptions~\cite{hendrycks2018benchmarking}, adversarial perturbation~\cite{szegedy2013intriguing}, and out-of-distribution examples~\cite{hendrycks2017a, hendrycks2021natural}.
Moreover, it is also known that image classification datasets contain spurious features~\cite{geirhos2020shortcut}, \pyone{which are associated features of the true class object and are detected by the classifiers.}

\begin{figure}[t]
\centering
\includegraphics[width=\linewidth]{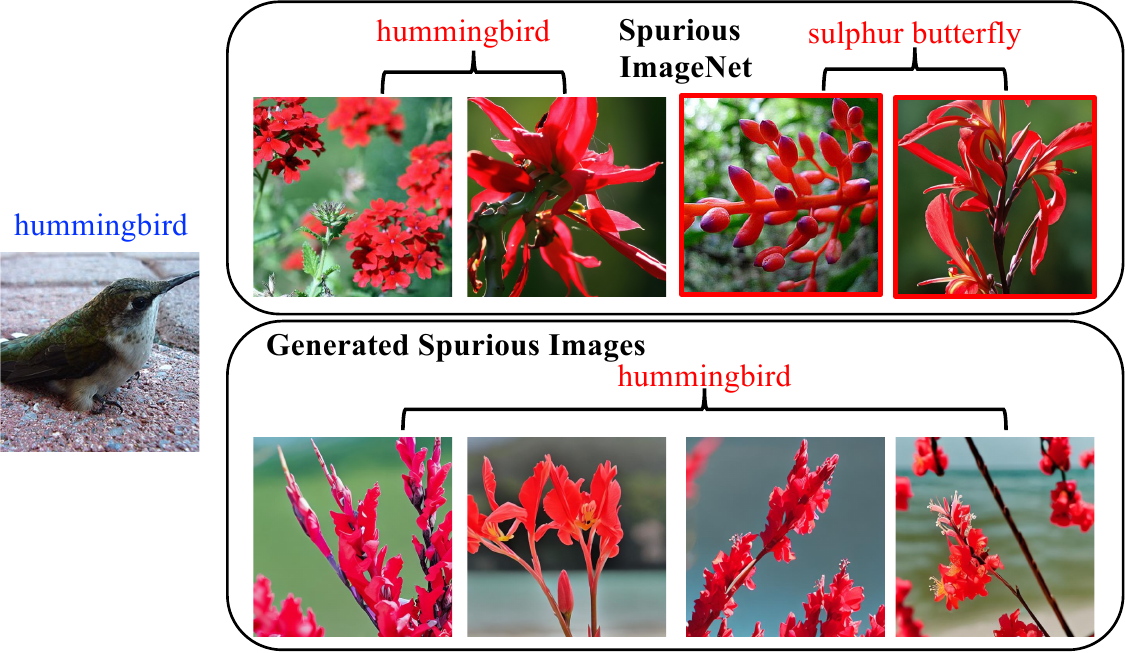}
\caption{Spurious images. Some images from \pyone{Spurious ImageNet~\cite{neuhaus2023spurious} dataset} are detected as ``hummingbird'' but classified as ``sulphur butterfly''.\label{fig:example}}
\end{figure}

Although researchers have put significant work into adversarial examples, there is less work on spurious features. Identifying and debugging image classifiers from spurious features is extremely important for safety-critical applications~\cite{adebayo2020debugging}. For example, a DNN-based pneumonia classifier performed well on internal data and not on external data due to hospital system-specific bias~\cite{zech2018variable}. A recent work detected spurious features from a large-scale dataset like ImageNet and introduced Spurious ImageNet~\cite{neuhaus2023spurious}. We found that not all images from Spurious ImageNet are spurious across different classifiers (Fig.~\ref{fig:example}). Moreover, filtering many images from the Internet for spurious features is time-consuming. Therefore, in this paper, we propose generating spurious images that are spurious across different classifiers by leveraging large-scale text-to-image models like Stable Diffusion~\cite{rombach2022high}, aiming to complement Spurious ImageNet. The advantage of using a generative model is that the generation has no upper bound.

We make the following contributions in this paper.
\begin{itemize}
  \item We propose a method for generating spurious images for the first time by fine-tuning Stable Diffusion with several reference spurious images from Spurious ImageNet. \pyone{The generated images by our method are useful for measuring the reliance on a classifier, and our method may be used to generate synthetic data for training a robust classifier.}
\item We introduce a new spurious feature similarity loss to distinguish between spurious and non-spurious images in Stable Diffusion's fine-tuning framework.
\item We conduct experiments to verify the effectiveness of the proposed method with analysis and discussion.
\end{itemize}
In experiments, the proposed method can generate new spurious images that resemble Spurious ImageNet, and generated spurious images are consistently spurious across different classifiers.

\section{Related Work}
\subsection{Spurious Features}
Spurious features coincide with an actual class object and are detected by a classifier. When only spurious features are associated with a class, they cause shortcut learning~\cite{geirhos2020shortcut}. For example, a cow is related to grass, and the classifier utilizes the grass to predict the cow. Therefore, a cow on the beach is not recognized because of the missing spurious feature grass~\cite{beery2018recognition}.

Spurious ImageNet defines spurious features in two settings: (1) spurious class extension and (2) spurious shared feature~\cite{neuhaus2023spurious}. Following the formulations and notations in~\cite{neuhaus2023spurious}, let $C_k$ be the set of all images under class $k$, $C_l$ be the set of all images under class $l$, and $S$ be the set of all images containing a feature, $s$. Spurious class extension occurs when a classifier detects $s$ for class $k$ even on $S \setminus C_k$ (\ie, the classifier predicts class $k$ even though it does not exist). A spurious shared feature occurs when two classes $C_k$ and $C_l$ share spurious features $s$ and the classifier predicts class $k$ for $S \cap C_l$ (\ie, the classifier favors the shared spurious feature $s$ over class $l$ and predicts class $k$ even though it does not exist).

Spurious ImageNet detects harmful spurious features by utilizing class-wise neural principal component analysis (NPCA) with minimal human supervision~\cite{neuhaus2023spurious}. Instead of filtering many images from the Internet, we aim to generate spurious features similar to Spurious ImageNet by leveraging text-to-image diffusion models in this paper.

\subsection{Text-to-Image Diffusion Models}
A diffusion model is a latent variable model that consists of a forward process (gradually adding Gaussian noise to a sample from a true data distribution) and a reverse process (gradually denoising Gaussian noise to a true data sample with learned Gaussian transitions~\cite{sohl2015deep,ho2020denoising}). Both processes are defined as Markov chains.

Diffusion models, specifically denoising diffusion probabilistic models (DDPM), have shown state-of-the-art results in image generation~\cite{dhariwal2021diffusion}. A text-to-image diffusion model is a diffusion model that can generate an image from an initial noise map and a text prompt such as Stable Diffusion~\cite{rombach2022high}, Imagen~\cite{saharia2022photorealistic}, GLIDE~\cite{pmlr-v162-nichol22a}, eDiff-I~\cite{balaji2022ediffi}, and DALL-E 2~\cite{ramesh2022hierarchical}. Large-scale models can also be adapted not only to text-to-image generation tasks but to many exciting downstream applications such as artistic painting~\cite{rombach2022text, avrahami2022blended}, spatial condition (sketch, segmentation map or layout) based image generation~\cite{zhang2023adding}, and image editing~\cite{brooks2023instructpix2pix, kim2022diffusionclip, mokady2023null, tsaban2023ledits}. In this paper, we use Stable Diffusion to generate spurious images.

\section{Method}
\begin{figure}[!t]
\centering
\includegraphics[width=\linewidth]{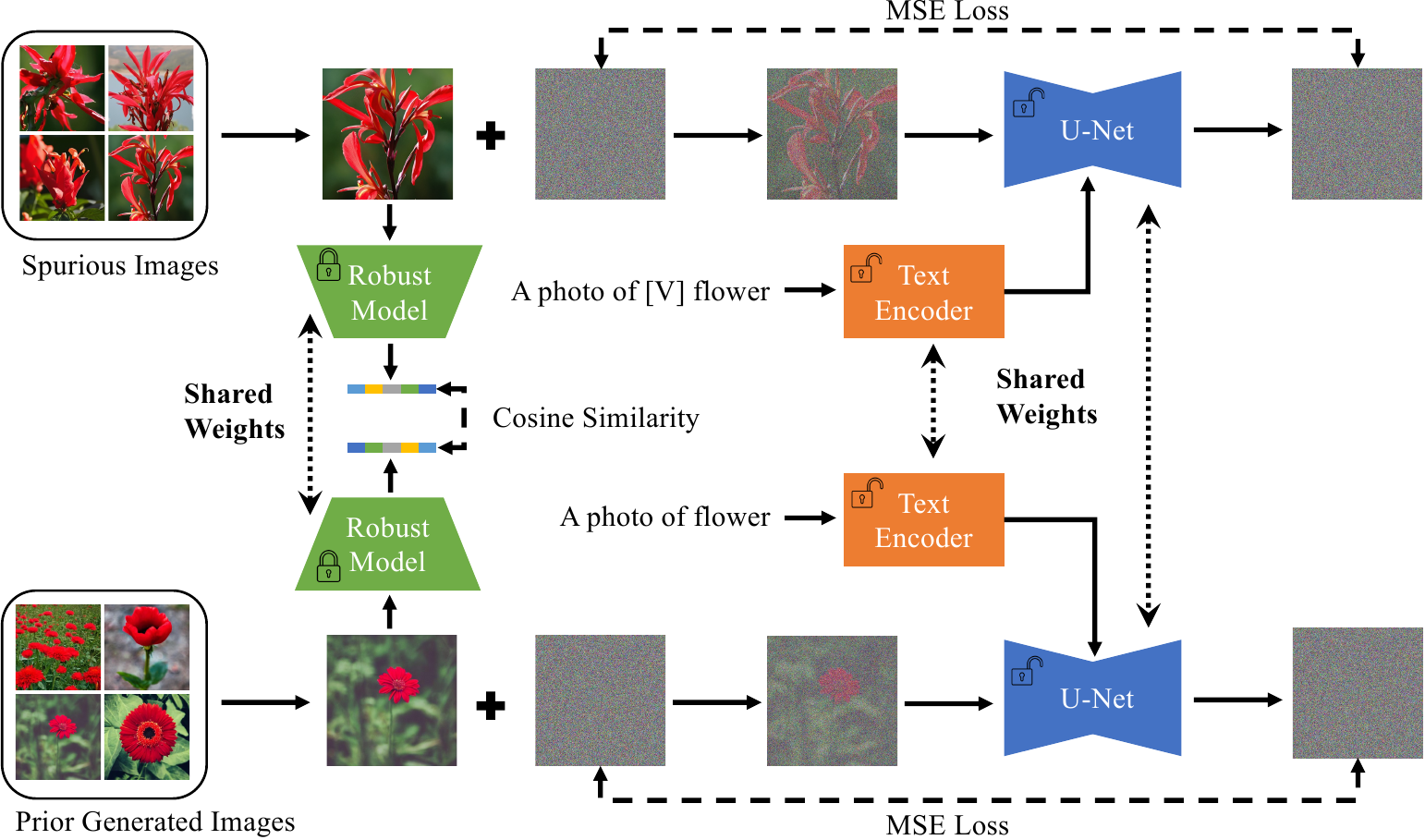}
\caption{Stable Diffusion fine-tuning. [V] indicates three-character unique identifier as in DreamBooth~\cite{ruiz2023dreambooth}.\label{fig:system}}
\end{figure}

Given a few spurious images of a class, we aim to generate new spurious images of this particular class that are spurious across different classifiers. Motivated by personalizing text-to-image diffusion models, we propose fine-tuning Stable Diffusion similar to DreamBooth~\cite{ruiz2023dreambooth} with a new spurious feature similarity loss. Figure~\ref{fig:system} illustrates the Stable Diffusion fine-tuning framework with the proposed loss. The significant differences between our method and DreamBooth are that we jointly fine-tune the text encoder and noise predictor and have an additional loss term in the Stable Diffusion fine-tuning objective. We next review Stable Diffusion and its personalization in detail and propose a new loss for distinguishing spurious features from non-spurious images to encourage spurious feature generation.

\subsection{Stable Diffusion}
Stable Diffusion~\cite{rombach2022high} is a large-scale text-to-image diffusion model widely known for its powerful generative ability and open-source release. To improve the efficiency and quality of diffusion models, Stable Diffusion works on the latent space of a pre-trained variational autoencoder with an encoder $\mathcal{E}$ and decoder $\mathcal{D}$. $\mathcal{E}$ encodes an image $x \in \mathbb{R}^{H \times W \times 3}$ in RGB space into a latent representation $z = \mathcal{E}(x)$, where $z \in \mathbb{R}^{h \times w \times c}$ such that $\mathcal{D}$ can decode $z$ as $\hat{x} = \mathcal{D}(z)$, which is visually identical to $x$. Basically, Stable Diffusion is a denoising diffusion probabilistic model (DDPM)~\cite{ho2020denoising} in latent space known as a latent diffusion model (LDM). Therefore, it learns a U-Net model, $\epsilon_\theta$, that predicts noise added to noisy latent $z_t$, which is latent $z = \mathcal{E}(x)$ with noise added at timestep $t \in T$. Given a text condition $y$ (\ie, text prompt), the objective of a text-to-image LDM is
\begin{equation}
 \mathcal{L_{\text{(LDM)}}} \coloneqq \mathbb{E}_{\mathcal{E}(x), y, \epsilon \sim \mathcal{N}(0, 1), t}[\lVert\epsilon - \epsilon_{\theta}(z_t, t, \tau_\theta(y))\rVert^2], \label{eq:ldm}
\end{equation}
where both $\epsilon_\theta$ and $\tau_\theta$ are jointly optimized. However, Stable Diffusion adopts a frozen CLIP text encoder instead of a trainable text encoder $\tau_\theta$.

\subsection{Personalizing Stable Diffusion}
Given a few images of a subject, the idea of personalization is to embed the subject into the output domain of Stable Diffusion to synthesize novel renditions of the subject in different contexts. There are two well-known techniques to achieve personalization.

One is to map a new text embedding $v_\ast$ corresponding to a new placeholder string $S_\ast$ and add $v_\ast$ to the text embedding space. This method is known as textual inversion. The new embedding $v_\ast$ is optimized by using a few reference images with the same LDM objective in~\eqref{eq:ldm} while keeping both text encoder $\tau_\theta$ and noise predictor $\epsilon_\theta$ frozen. It works well for an LDM in which $\tau_\theta$ and $\epsilon_\theta$ are jointly trained.

Another technique is to carefully fine-tune a large text-to-image model like Stable Diffusion to integrate new information on a subject into the output domain without overfitting it to a small number of reference images or forgetting prior knowledge. This technique is known as DreamBooth~\cite{ruiz2023dreambooth}, which fine-tunes the noise predictor $\epsilon_\theta$ with reference images and text prompts that contain a unique identifier (\eg, a photo of an [identifier] flower). To retain prior knowledge, DreamBooth introduces class-specific prior preservation loss (PPL), \ie,
\begin{equation}
\mathcal{L_{\text{(PPL)}}} \coloneqq \mathbb{E}_{\mathcal{E}(x'), y', \epsilon' \sim \mathcal{N}(0, 1), t}[\lVert\epsilon' - \epsilon_{\theta}(z'_t, t, \tau_\theta(y'))\rVert^2],
\end{equation}
where $x'$ is generated from pre-trained Stable Diffusion with text prompts (a photo of [class]) without the [identifier]. The new objective for DreamBooth in Stable Diffusion is
\begin{equation}
\mathcal{L_{\text{(DreamBooth)}}} \coloneqq \mathcal{L_{\text{(LDM)}}} + \lambda \mathcal{L_{\text{(PPL)}}},
\end{equation}
where $\lambda$ is a hyperparameter. We build the proposed method on DreamBooth and introduce a new spurious feature similarity loss to enforce spurious feature generation from Stable Diffusion.

\subsection{Spurious Feature Similarity Loss}
We propose spurious feature similarity loss (SFSL) inspired by Spurious ImageNet to encourage spurious feature generation. Spurious ImageNet~\cite{neuhaus2023spurious} uses an adversarially robust model to detect spurious features because prior works showed that robust models have generative properties~\cite{santurkar2019image, tsipras2018robustness}. Specifically, Spurious ImageNet considers features of the penultimate layer $\phi(x) \in \mathbb{R}^D$ of an improved adversarially trained model from~\cite{croce2022adversarial}. Given a class $k$, a class-wise feature of a robust model is defined in Spurious ImageNet as
\begin{equation}
\psi_k(x) = w_k \odot \phi(x),
\end{equation}
where $w_k \in \mathbb{R}^D$ is class-wise weights of the final layer. Principal component analysis (PCA) is applied to such class-wise neural features to detect spurious features in Spurious ImageNet. We observe that class-wise neural features are highly similar among spurious images. On the basis of this, we introduce SFSL by calculating cosine similarity $S_C$ between class-wise neural features of prior generated images, as in DreamBooth, and those of spurious reference images as
\begin{equation}
\mathcal{L_{\text{(SFSL)}}} \coloneqq S_C(\psi_k(x), \psi_k(x')).
\end{equation}
Our new objective for fine-tuning Stable Diffusion is
\begin{equation}
\mathcal{L_{\text{(Proposed)}}} \coloneqq \mathcal{L_{\text{(LDM)}}} + \lambda \mathcal{L_{\text{(PPL)}}} + \kappa \mathcal{L_{\text{(SFSL)}}},
\end{equation}
where $\kappa$ is a hyperparameter, and we jointly train both $\tau_\theta$ and $\epsilon_\theta$.

\section{Experiments}
To verify the proposed method's effectiveness, we conducted experiments by fine-tuning Stable Diffusion for six classes from Spurious ImageNet~\cite{neuhaus2023spurious}: hummingbird, freight car, fireboat, gondola, flagpole, and koala.

\subsection{Setup}
\noindent{\bf Dataset.} We utilized Spurious ImageNet~\cite{neuhaus2023spurious}, which contains 100 classes. For each class, there are 75 spurious images with a resolution of $367 \times 367$, totaling 7,500 images. We found that all images from Spurious ImageNet are not consistently spurious across different classifiers. Therefore, we selected six images for each test class that are all spurious for four classifiers: ResNet-50~\cite{he2016deep} from the PyTorch library\footnote{\scriptsize \url{https://pytorch.org/vision/main/models/generated/torchvision.models.resnet50.html\#torchvision.models.ResNet50\_Weights}} with weights version 1 and 2, robust ResNet-50~\cite{croce2022adversarial}, and ViT-B/16 with augmentation and regularization~\cite{steiner2022how}.

\vspace{2mm}
\noindent{\bf Implementation.} We implemented the proposed method on top of DreamBooth~\cite{ruiz2023dreambooth} with the proposed additional loss term by using the HuggingFace diffuser library\footnote{\scriptsize \url{https://github.com/huggingface/diffusers}}. In our training, we used pre-trained Stable Diffusion v1.5\footnote{\scriptsize \url{https://huggingface.co/runwayml/stable-diffusion-v1-5}}. We used the AdamW optimizer~\cite{loshchilov2018decoupled} with a decay value of $0.01$, a beta1 value of 0.9, a beta2 value of 0.999, an epsilon value of $1\times e^{-8}$, a fixed learning rate value of $2\times e^{-6}$, and a batch size value of 1. As in DreamBooth, we used a prior preservation loss weight with a value of 1. We also set the spurious similarity loss weight to 1. We fine-tuned U-Net and the text encoder in Stable Diffusion for 800 steps using the six filtered spurious images for each test class. For evaluation, we used PNDMScheduler~\cite{liu2022pseudo} fixed latents with the same seeds and sampled the fine-tuned Stable Diffusion for 25 steps with a classifier-free guidance scale value of $7.5$.

\subsection{Spurious Accuracy}
We sampled 75 images for each test class and observed the spurious class accuracy across the four classifiers, ResNet-50 V1 and V2, Robust ResNet-50, and ViT-B/16, as described above, while comparing with Spurious ImageNet. The 75 generated images were non-cherry-picked. Table~\ref{tab:sp-accuracy} summarizes the spurious accuracy results, where SI denotes Spurious ImageNet. In all test classes except for ``flagpole,'' the generated images were more spurious across different classifiers. This proves that the proposed method is complementary to Spurious ImageNet in evaluating existing ImageNet classifiers for spurious performance. Using the proposed method, one can design a more robust spurious test dataset.

\begin{table*}
 \centering
 \resizebox{\linewidth}{!}{%
 \begin{tabular}{lcccccccccccc} %@{} surpasses the leading space between columns
 \toprule
 \multirow{2}{*}{\backslashbox{Model}{Dataset}} & \multicolumn{2}{c}{Hummingbird} & \multicolumn{2}{c}{Freight Car} & \multicolumn{2}{c}{Fireboat} & \multicolumn{2}{c}{Gondola} & \multicolumn{2}{c}{Flagpole} & \multicolumn{2}{c}{Koala}\\
 \cmidrule{2-13}
 & SI & Ours & SI & Ours & SI & Ours & SI & Ours & SI & Ours & SI & Ours\\
 \midrule
 ResNet-50~\cite{he2016deep} (V1) & 60.00 & \bf 90.67 & 73.33 & \bf 94.67 & 66.67 & \bf 81.30 & 49.33 & \bf 100.00 & 94.67 & \bf 100.00 & 48.00 & \bf 52.00\\
 ResNet-50~\cite{he2016deep} (V2) & 37.33 & \bf 60.00 & 89.33 & \bf 100.00 & 68.00 & \bf 85.33 & 73.33 & \bf 100.00 & 97.33 & \bf 98.67 & 50.67 & \bf 56.00\\
 ResNet-50 (Robust)~\cite{croce2022adversarial} & 93.33 & \bf 100.00 & 98.67 & \bf 100.00 & 64.00 & \bf 82.67 & 98.67 & \bf 100.00 & \bf 100.00 & 98.67 & 48.00 & \bf 66.67\\
 ViT-B AugReg~\cite{steiner2022how} & 33.33 & \bf 98.00 & 82.67 & \bf 96.00 & 45.33 & \bf 80.00 & 50.67 & \bf 100.00 & 97.33 & \bf 98.67 & 38.67 & \bf 78.67\\
 \bottomrule
 \end{tabular}
 }
\caption{Spurious accuracy ($\%$) of generated images and Spurious ImageNet (SI)~\cite{neuhaus2023spurious}. V1 and V2 indicate corresponding weight versions.\label{tab:sp-accuracy}}
\end{table*}

\subsection{Comparison}
To the best of our knowledge, we are the first to generate spurious images by leveraging large text-to-image models. Therefore, there is no method for a direct comparison. Nevertheless, as we built on DreamBooth~\cite{ruiz2023dreambooth}, we compared our method with the vanilla DreamBooth by carrying out an ablation study. Table~\ref{tab:ablation} compares the average spurious accuracy of the six classes across the four classifiers. The generated images were made more spurious by jointly training the text encoder with DreamBooth. The spurious accuracy was further increased when adding the proposed spurious features similarity loss (SFSL). We observed that the hyperparameter $\kappa$ has a different effect depending on the class. A $\kappa$ value of 0.8 produced better spurious images for three test classes: freight car, gondola, and koala. On the other hand, a $\kappa$ value of 1 was better for the other test classes. Therefore, we recommend choosing the $\kappa$ value on the basis of the target class.

\begin{table}
\centering
\caption{Ablation study.\label{tab:ablation}}
\resizebox{\columnwidth}{!}{%
\begin{tabular}{lc}
 \toprule
 {Ablation} & {Average Spurious Accuracy ($\%$) $\uparrow$}\\
 \midrule
 {Vanilla DreamBooth~\cite{ruiz2023dreambooth}} & 69.06\\
 {+ Trainable text encoder} & 91.17 \\
 \midrule
 {+ SFSL ($\kappa = 1$)} & 88.25\\
 {+ SFSL ($\kappa = 0.8$)} & \bf 93.83\\
 {+ SFSL ($\kappa = 0.5$)} & 83.61\\
 \bottomrule
\end{tabular}
}
\end{table}

\subsection{Perceptual Quality}
We utilized the latest perceptual image quality assessment metric, TOPIQ~\cite{chen2023topiq} (non-reference version), to objectively measure the perceptual quality of generated spurious images.
Table~\ref{tab:topiq} summarizes the objective evaluation results where the TOPIQ score is calculated for 6 images for each class (all training images) and 75 images for each class (generated ones).
The scores of generated images for ``hummingbird'' and ``koala'' were close to that of real ones.
However, the scores were lower for the other classes.
To further evaluate the quality of spurious generated images, we carried out the subjective evaluation in the next sub-section.

\begin{table}
\centering
\caption{Average TOPIQ~\cite{chen2023topiq} score of $n$ images.\label{tab:topiq}}
% \resizebox{\columnwidth}{!}{%
\begin{tabular}{lcc}
 \toprule
 \multirow{2}{*}{Class} & \multicolumn{2}{c}{TOPIQ~\cite{chen2023topiq} ($\%$) $\uparrow$}\\
                        & {Real ($n = 6$)} & {Generated ($n = 75$)}\\
 \midrule
 {Hummingbird} & 0.62 & \underline{0.60}\\
 {Freight Car} & 0.65 & 0.49\\
 {Fireboat} & 0.55 & 0.37\\
 {Gondola} & 0.69 & 0.45\\
 {Flagpole} & 0.63 & 0.47\\
 {Koala} & 0.66 & \underline{0.56}\\
 \bottomrule
\end{tabular}
% }
\end{table}

\subsection{Subjective Evaluation}
\begin{figure}[t]
\centering
\includegraphics[width=\linewidth]{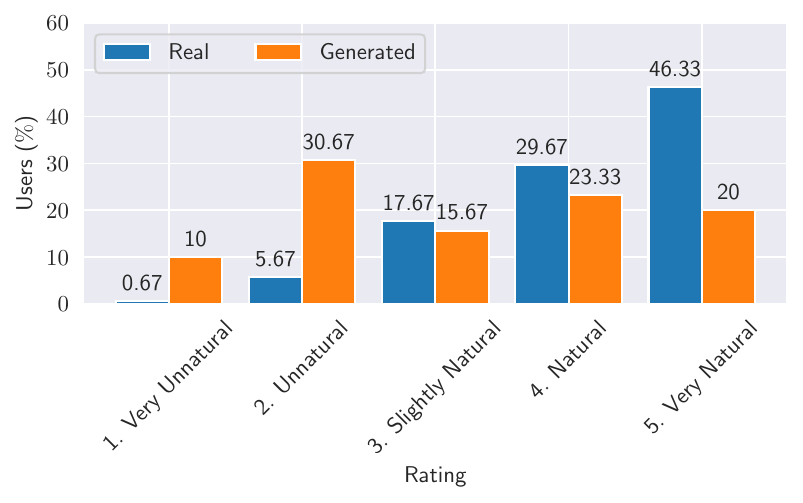}
\caption{Subjective evaluation of real and generated images.\label{fig:subjective}}
\end{figure}

\begin{figure}[htb]
\centering
\includegraphics[width=\linewidth]{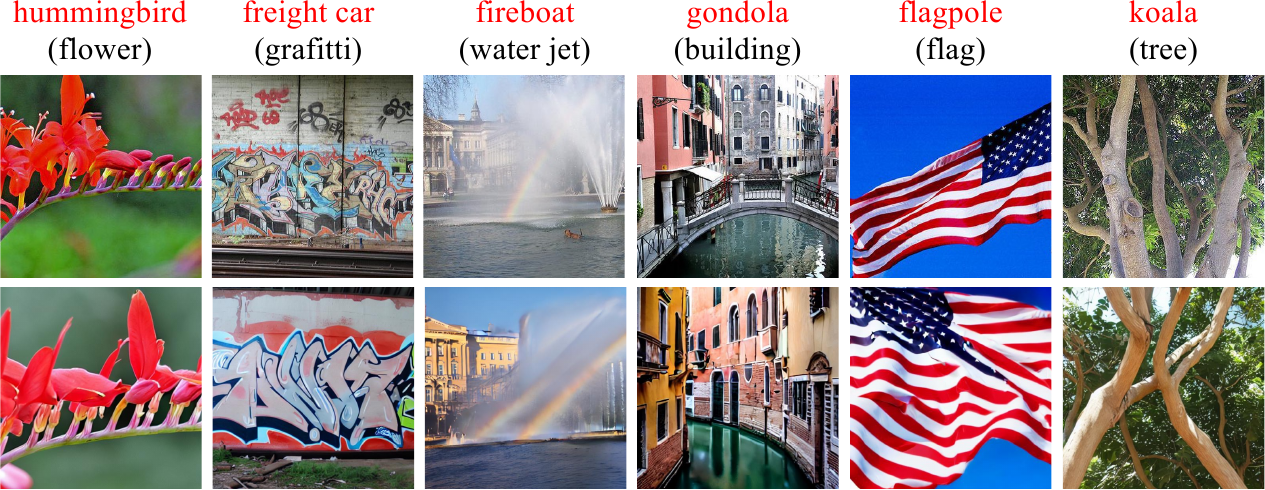}
\caption{Selected examples of generated images (second row) and Spurious ImageNet (first row). Red label describes predicted class, and black label is true subject.\label{fig:generated}}
\end{figure}

We conducted a subjective evaluation with 10 users, including researchers, students, and non-technical people.
We showed users 10 random images for each class  (a mixture of real and generated ones) and asked them to rate from 1 to 5 according to naturalness.
Figure~\ref{fig:subjective} summarizes subjective evaluation results.
On average, $46.33\%$ of users gave the highest score, 5 (very natural), to real images and $20\%$ to the generated ones.
The results show that some of the generated images are natural and realistic.
In addition, we manually checked the generated images of all six classes.
We observed diffusion artifacts in some images.
Figure~\ref{fig:generated} shows selected generated images compared with images from Spurious ImageNet. 
However, as generative models do not have an upper bound of generation, one can sample many images with different configurations, such as different starting noise and guidance scales, to obtain satisfiable images.

\subsection{Recontextualization}
\begin{figure}[htb]
\centering
\includegraphics[width=\linewidth]{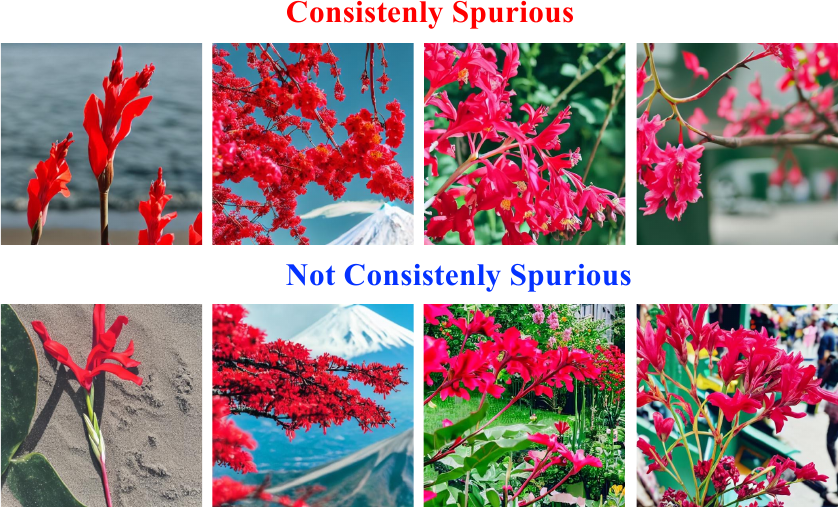}
\caption{Recontextualized spurious images. Left to right images were generated with prompts, \emph{``a photo of a [V] flower on the beach, $\ldots$ on Mount Fuji, $\ldots$ in a garden, $\ldots$ in a market''}. Images in first row are classified as ``hummingbird,'' and those in second row are not.\label{fig:new-context}}
\end{figure}

We carried out experiments on generating spurious features in different environments. Figure~\ref{fig:new-context} shows such recontextualized images. When a new context is visible in generated images, spurious features are not consistently classified as spurious across different classifiers. From the figure, we can see that a new contextualized background is blurred in consistently spurious images, while spurious images with new scenes are only spurious to one classifier or not spurious at all. This may be due to fine-tuning the text encoder in the proposed method. In our future work, we shall further investigate how spurious strength changes in different contexts. In addition, we shall also experiment with fine-tuning the text encoder with an efficient fine-tuning technique, LoRA~\cite{hu2022lora}, to preserve the prior knowledge of the text encoder.

\section{Discussion}
We demonstrated a fine-tuning approach that adds an additional constraint to DreamBooth fine-tuning to generate spurious images by using several reference images from Spurious ImageNet. While our method can produce spurious images resembling reference spurious ones, we inherit the limitations of Stable Diffusion and its fine-tuning techniques. For example, Stable Diffusion-generated images may contain unwanted artifacts, such as humans with strange appearances, and DreamBooth may overfit the fine-tuned images.

\begin{figure}[htb]
\centering
\includegraphics[width=\linewidth]{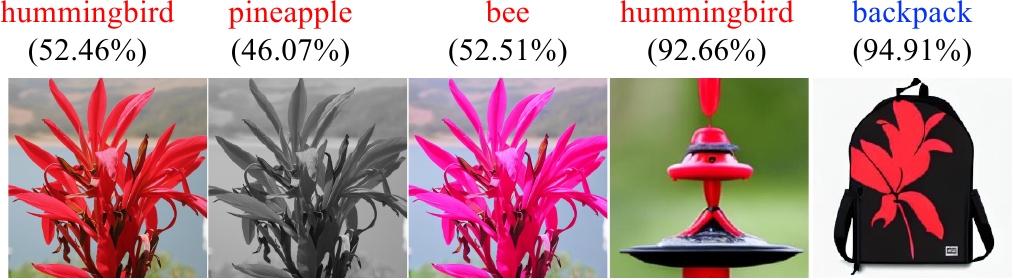}
\caption{Shape and style bias of spurious image.\label{fig:analysis}}
\end{figure}

Although it has been pointed out that the spurious features of Spurious ImageNet are context-dependent and exist due to a bias in the dataset, we are interested in knowing what makes spurious features spurious. We ran a simple experiment by changing the shape and style of a spurious image. Figure~\ref{fig:analysis} shows a spurious feature that heavily depends on the subject's color. A red flower is classified as a hummingbird. When the color of the flower is changed, the model's decision is also changed accordingly, even though it is the same flower. Moreover, a red bird feeder is also classified as a hummingbird. However, when we transfer the style of the red spurious flower to a backpack, the spurious feature is not spurious anymore. Therefore, a spurious feature is context-dependent and biased towards shape and style. This raises whether a robust spurious feature can work as a backdoor (context-independent spurious feature).

Another interesting point is whether a spurious feature can be a class itself. Spurious ImageNet and our method focus on a spurious feature without a class object. However, a classifier may always prefer a spurious class, even though it might appear trivial in a given scene. Further research is required to understand spurious features in a large-scale dataset like ImageNet.

\section{Conclusion}
In this paper, we demonstrate that given a few spurious images from Spurious ImageNet, we can fine-tune Stable Diffusion by leveraging a new spurious feature similarity loss to generate spurious images. Our proposed method can bypass the time-consuming filtering of many images to find spurious features. Thus, the proposed method complements Spurious ImageNet in preparing a spurious feature testing dataset. Experiments confirmed that generated images are spurious across different classifiers and are visually similar to images in Spurious ImageNet.


\begin{small}
\bibliographystyle{IEEEbib}
\bibliography{refs}
\end{small}

\end{document}